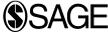

# Fast and computationally efficient generative adversarial network algorithm for unmanned aerial vehicle–based network coverage optimization

Marek Ružička[1], Marcel Vološin[1], Juraj Gazda[1], Taras Maksymyuk[2], Longzhe Han[3] and MisCha Dohler[4]

## Abstract
The challenge of dynamic traffic demand in mobile networks is tackled by moving cells based on unmanned aerial vehicles. Considering the tremendous potential of unmanned aerial vehicles in the future, we propose a new heuristic algorithm for coverage optimization. The proposed algorithm is implemented based on a conditional generative adversarial neural network, with a unique multilayer sum-pooling loss function. To assess the performance of the proposed approach, we compare it with the optimal core-set algorithm and quasi-optimal spiral algorithm. Simulation results show that the proposed approach converges to the quasi-optimal solution with a negligible difference from the global optimum while maintaining a quadratic complexity regardless of the number of users.

## Introduction

We are rapidly developing a digitized, highly connected, and data-driven society, where most of the business and industrial operations and citizen's daily routines rely on ubiquitous wireless connectivity. To date, the growing demand for data-hungry mobile applications has forced operators to deploy many additional small cells to accommodate the ever-increasing traffic volumes. However, the gains achieved by the small cells are very sensitive to the mobility pattern of user equipments (UEs). Typically, each city has its own unique social and economic features, which affect the movement of citizens and are considered by mobile operators when deploying small cells in the most effective way. However, there are also various exogenous factors that can temporary change the typical mobility patterns, such as big social events or sports tournaments. The most recent example is the coronavirus

[1]Department of Computers and Informatics, Technical University of Košice, Košice, Slovakia
[2]Department of Telecommunications, Lviv Polytechnic National University, Lviv, Ukraine
[3]School of Information Engineering, Nanchang Institute of Technology, Nanchang, China
[4]Centre of Telecommunications Research, King's College London, London, UK

**Corresponding author:**
Juraj Gazda, Department of Computers and Informatics, Technical University of Košice, Letna 9, Košice 04001, Slovakia.
Email: juraj.gazda@tuke.sk

disease 2019 (COVID-19) pandemic, which has caused a massive change in the lifestyle and typical mobility patterns of people worldwide due to massive lockdowns and quarantine restrictions. According to the Google Mobility Trends Report, in 2020, the activity of UEs in business areas has dropped by 30%, while the activity in residential areas and parks has increased by 20% and 109%, respectively.[1] Such challenges of abrupt changes in UEs mobility can be effectively solved by moving cells based on unmanned aerial vehicles (UAVs).

Recently, UAV-based wireless networks have been gaining momentum for various applications such as disaster recovery, telemetry, or military communications.[2] However, to be used effectively as an alternative to the cellular mobile network infrastructure, UAVs need to be optimally placed within the target coverage area. There are many proposed solutions for the deployment of UAV-based communication systems, which are covered in recent surveys and research articles.[3–9]

In this article, we propose a novel solution for optimal coverage design based on a conditional generative adversarial network (cGAN). cGANs belong to the class of generative deep learning models, which are widely adopted for various challenges in fifth-generation (5G) systems.[10,11] The basic motivation for using a cGAN is its ability to learn a mapping from input data to a target data distribution. While most of the cGAN research targets image or video generation, there is tremendous potential for cGAN technology in other fields of computer science, such as clustering[12] and intrusion detection.[13] In our contribution, the input data are represented by the UEs distribution in the space, and the target data are represented by the UAVs distribution providing the coverage for UEs. The computation of the optimal UAVs distribution is generally NP-hard[9] and thus, due to the extreme time limitations, not feasible in practice. We believe that through the fast and computationally efficient cGAN application, we can provide a sufficiently accurate approximation of the distributions of UAVs with a reduced time complexity. As the baseline for comparison, we chose two algorithms, which provide either the optimal solution for the UAV-based coverage, termed the core-set algorithm,[7,8] or a quasi-optimal solution, termed the spiral algorithm.[9] Note that in the article, we do not target the specific mobility of UAVs but focus our attention on the computation of the UAVs positions for a given snapshot of UEs positions. Nevertheless, an important characteristic is that our algorithm is capable of *cooperating* with specific algorithms that address the mobility patterns of UAVs (e.g. based on the application of reinforcement learning (RL)) to achieve joint superior performance even in very complex and dynamic scenarios.

The remainder of this article is organized as follows. The *Related work* section briefly summarizes the recent advancements in UAV coverage optimization. In the *System model* section, we briefly describe the setup of the system model, basic assumptions, and constraints. In the *Proposed approach* section, we provide a detailed explanation of the proposed approach and implementation of the cGAN workflow for UAV-based coverage optimization. We split this section into two main subsections covering the *training phase* and the *deployment phase*. In the *Performance evaluation and discussion* section, we simulate and discuss the performance of the proposed approach. Finally, we conclude the article in the *Conclusion* section.

## Related work

The problem of optimal placement of UAVs can be formulated as a mixed integer nonlinear problem, which is generally NP-hard. Thus, we seek approximate solutions based on the application of either differential evolution strategies or machine learning algorithms. Among the former methods, we can appreciate the contribution of Plachy et al.,[14] where the authors elaborated on the application of particle swarm optimization (PSO) for UAVs placement, reflecting the instantaneous positions of UEs. Later, the model was extended by including the interference between the base stations and UEs by leveraging the fundamental electricity forces and well-known Coulomb's law.[15] The optimal path of UAVs that maintains a fixed operational altitude and avoids obstacles was provided by Gonzales et al.[16] The authors suggested the application of the differential evolution concept for a feasible UAV path. The proposed approach is of special relevance to 5G/sixth-generation (6G) communication systems, as it deals with different vision fields of the UAV, which is very common in 6G communication systems supported by UAVs in urban areas. Finally, self-organization of a large fleet of UAVs that follows the specific mobility pattern of UEs was recently proposed in Horváth et al.[17] The authors employ an evolutionary strategy based on the existence of virtual forces and show the superiority of the proposed approach compared to the conventional rule-based methods.

Machine learning algorithms for UAV coverage mainly rely on the application of unsupervised clustering algorithms[9] and RL methods. A promising solution in this domain is the combination of both: deep neural nets (NNs) for large state space approximation and RL rules applied in the discrete-time stochastic process controlled by a Markov decision process (MDP). The advantage of a deep RL-based approach is that it captures the system evolution trajectory and does not require any tuning in the validation phase once the control policy is trained. Pioneering works addressing UAVs coverage optimization based on the application of deep RL were proposed by Wang et al. and Cui et al.[18,19] The authors elaborated on the application of a conventional single-agent RL strategy represented by the state-action-reward-state-action (SARSA) algorithm,[18] and the concept was also extended to the multiagent learning scenario, which is of more relevance in the presence of a large fleet of UAVs.[19] Furthermore, Khan and Yau[20] proposed the application of deep RL in determining the optimal UAV trajectory while addressing the limited energy resources of UAVs, and thus, optimization criteria were proposed for when the target was to increase the network lifetime of the UAV fleet. Finally, the work presented in Liu et al.[21] proposes a multiagent greedy-model-based RL approach that accounts for multiple UAVs with different parameters to explore environments in parallel to accelerate training.

In our work, we aim to propose a solution based on a cGAN application that could act as a *complementary mechanism* to the algorithms given above. As long as there exists high uncertainty related to the UAV position in the fleet, the proposed algorithms can only act weakly, providing substantial performance gaps compared to the maximum theoretical expectations. For example, RL-based algorithms determine the positions of UAVs in fleet indirectly by proposing a specific reward function, which cannot generalize well in complex state spaces. Hence, our proposed approach intends to implement computation of quasi-optimal locations of UAVs in a fleet for a given snapshot of UEs positions in a very short time frame compared to the state-of-the-art algorithms (e.g. the core-set algorithm and spiral algorithm). While both algorithms are characterized by high time complexity that limits their application in the practice, we believe that the low time complexity of our proposed approach is of paramount importance. Thus, our proposed low-complexity approach (*computation of the UAVs positions in the fleet*) can co-exist and cooperate with the complex deep RL approach (*computation of the trajectories of individual UAVs in the fleet*) to jointly achieve truly high-precision and low-complexity solutions, targeting the theoretically achievable upper-bound performance of UAV-based communication systems. Note that we only address the former problem (*computation of the UAVs positions in the fleet*) throughout the article, and the combination of deep RL and our proposed approach is left for our future work.

## System model

To perform coverage optimization, we split the investigated area into nonoverlapping segments of size $n \times n$. Thus, we represent the scenario in matrix form, where each matrix element represents the number of UEs in a segment and their corresponding locations in the area. The same procedure is applied to the UAV-based aerial base stations. Segmentation of the investigated region allows application of powerful deep learning techniques and fast matrix processing to determine the quasi-optimal coverage for UEs. Here, we assume that the UE is covered if it is positioned within the dedicated perimeter of the UAV. Our proposed approach is aimed at determining the minimum required number of UAVs and their corresponding positions to attain the quasi-optimal coverage of the target area using a cGAN.

Throughout the article, we use the following notations. The matrix $\mathbf{X} = (x_{ij}) \in \mathcal{N}^{n \times n}$ represents the distribution of UEs in the area, while $x_{ij}$ denotes the number of UEs in the $(i,j)$ segment. Similarly, $\mathbf{K} = (k_{ij}) \in \mathcal{N}^{n \times n}$ and $\mathbf{Y} = (y_{ij}) \in \mathcal{N}^{n \times n}$ denote the UAVs placement in the segmented region computed using the analytical core-set strategy and heuristic-based cGAN, respectively.

## Proposed approach

The proposed approach consists of two phases—the *training phase* of the cGAN and the *deployment phase*. The training phase encompasses the training of both architectures of the cGAN: generator $G$ and discriminator $D$. The deployment phase includes the joint application of the generator $G$ and a unique correction procedure avoiding blurring of the UAVs positions. The training is executed only once and *offline*; thus, it does not affect the time complexity of the proposed approach in practical deployments.

### Training phase

In this section, we introduce the assumptions considered throughout the article, cGAN model training/characteristics and unique multilayer sum-pooling loss that extends the conventional cGAN loss definition.

*Definitions and assumptions.* The cGAN[22] uses two deep neural networks simultaneously, that is, the generator ($G$) and discriminator ($D$). The training process of the cGAN proceeds by alternately updating the parameters of $G$ and $D$. In conventional GAN models,[23] the generator's task is to generate realistic samples, and the discriminator's task is to distinguish synthetic (generated) samples from real samples. For our problem, we use a cGAN, which allows adding an external condition of the spatial UEs distribution $\mathbf{X}$ in the coverage area for both the generator and the discriminator. Thus, a target objective of the generator $G$ is to compute the required minimum number of UAVs and their quasi-optimal locations $\mathbf{Y}$ for the given cGAN input. The role of the discriminator $D$ is to decide whether the generated positions of UAVs are produced by the optimal core-set algorithm or by the heuristics-based generator. In this case, the optimal locations of UAVs represent positions such that all UEs are covered with the minimal number of UAVs. We compute the optimal solution $\mathbf{K}$ based on the Euclidean $k$-center problem employing the *mixed breadth-depth traversing strategy*. In general, the solution referred to as the core-set algorithm is based on the existence of a small subset of points called *core-sets*, which results in a polynomial time approximation scheme for a fixed number of UAVs.

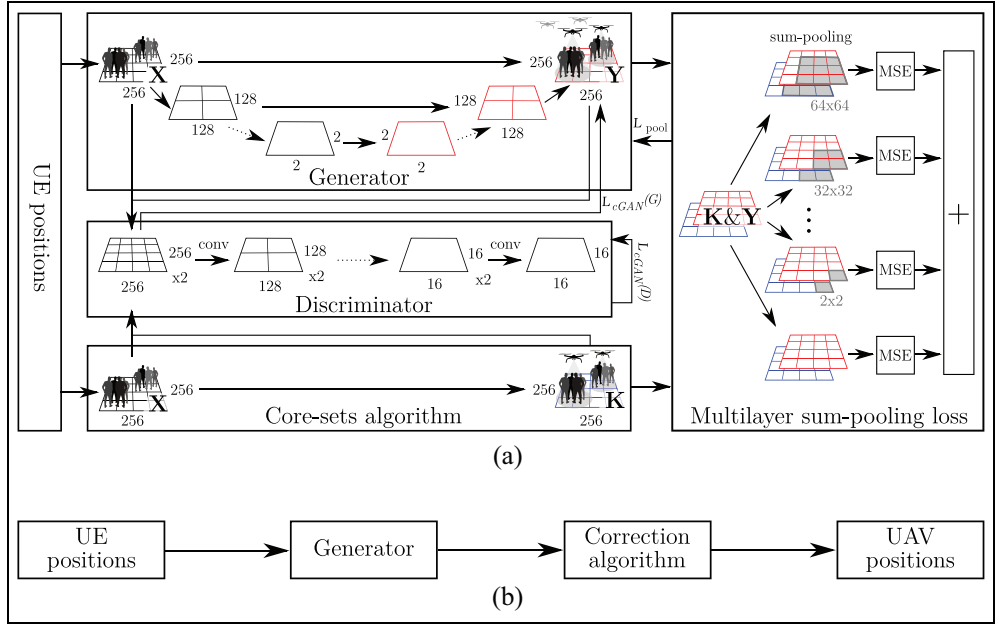

**Figure 1.** (a) Detailed block scheme of the training phase of the proposed approach. Note that the training phase is executed *offline* prior to the practical deployment of UAVs. The UAVs positions generated by the *core-set* algorithm serve as the template for our approach, determining the structure (weights and biases) of the generator G and discriminator D of the cGAN. In addition, a graphical illustration of the multilayer sum-pooling loss is also introduced. (b) Block scheme of the deployment phase executed *online*. The deployment phase consists of the trained generator G and the correction algorithm that avoids UAV position blurring.

*cGAN model.* We represent the coverage optimization problem in matrix form. Formally, the generator $G$ learns a mapping from input data $\mathbf{X}^{n \times n}$ to generate output data $\mathbf{Y}^{n \times n}$, $G : \mathbf{X} \to \mathbf{Y}$. The real data samples (optimal UAVs positions) computed by the core-set algorithm are denoted by the template matrix $\mathbf{K}^{n \times n}$.

The objective function of the generator $G$ aims to minimize the probability that the generated samples $G(\mathbf{X}) \to \mathbf{Y}$ are classified by the discriminator as synthetic. Thus, the objective function is presented in the following form

$$\min_{G} \mathcal{L}_{cGAN}(G) = \mathbb{E}_{\mathbf{Y}}[\log(1 - D(\mathbf{X}, \mathbf{Y}))] \quad (1)$$

The discriminator $D$ outputs a single scalar value, which represents the probability that $\mathbf{Y}$ is a sample produced by the generator $G$ rather than a sample from the template matrix $\mathbf{K}$. The objective function of the discriminator $D$ is defined as

$$\max_{D} \mathcal{L}_{cGAN}(D) = \mathbb{E}_{\mathbf{X}, \mathbf{K}}[\log D(\mathbf{X}, \mathbf{K})] \\ + \mathbb{E}_{\mathbf{Y}}[\log(1 - D(\mathbf{X}, \mathbf{Y}))] \quad (2)$$

Both the discriminator $D$ and generator $G$ are trained in parallel with the aim to minimize $\mathcal{L}_{cGAN}(G)$ and at the same time maximize $\mathcal{L}_{cGAN}(D)$. Isola et al.[24] further extended the training process with the L1 norm (Manhattan distance, i.e., mean absolute error (MAE)). The L1 norm is defined as the MAE between target $\mathbf{K}$ and generated data $\mathbf{Y}$. Due to its nature, we can formulate the iterative optimization problem as a minimax problem over $\mathcal{L}_{cGAN}(G, D)$ as follows

$$\arg \min_{G} \max_{D} \mathcal{L}_{cGAN}(G, D) + \lambda \mathcal{L}_{L1}(G) \quad (3)$$

where $\lambda = 100$ is the weighting coefficient.[24]

*Generator and discriminator.* The generator $G$ and discriminator $D$ are implemented according to the scheme shown in Figure 1(a), which depicts the training process

**Algorithm 1.** Multilayer sum-pooling loss

1: Inputs: $\mathbf{Y}^{n \times n}$, $\mathbf{K}^{n \times n}$
2: Initialize the loss matrix $\mathcal{L}_{pool} = \emptyset$
3: Set the sum-pooling filter size to $f = 1$, $stride = 1$, and $padding = 1$
4: **while** $f \leqslant 64$ **do**
5: Calculate the pooled version of the optimal position matrix with filter size $f$: $P_1 = pool(\mathbf{K}, f, padding)$
6: Calculate the pooled version of the generated position matrix with filter size $f$: $P_2 = pool(\mathbf{Y}, f, padding)$
7: Calculate the squared error for the pooled matrices $Loss = (P_1 - P_2)^2$
8: Add the loss to the total loss $\mathcal{L}_{pool} = \mathcal{L}_{pool} + Loss$
9: $f = 2 \times f$, $padding = 2 \times padding$
10: Output: $\mathcal{L}_{pool}$

for the proposed cGAN-based solution. For brevity, we refer to the detailed descriptions of the generator $G$ and discriminator $D$ in Isola et al.[24] In particular, generator $G$ is implemented as the U-net with an input shape of $256 \times 256$, according to the initial segmentation of the target area. The shape of each successive layer is decreased twice in both rows and columns, with the eventual bottleneck of $2 \times 2$ in the seventh layer. The discriminator $D$ is implemented as a conventional convolutional neural network, with the respective input shape of $256 \times 256$.

*Multilayer sum-pooling loss.* Since UAVs occupy only a few segments in the investigated area, the target matrix $\mathbf{Y}$ is extremely sparse. Therefore, application of the conventional loss function for a cGAN such as the L1 norm, as shown in equation (3), results in weak performance of the proposed approach.

To overcome this problem, we propose a custom multilayer sum-pooling loss function, which is computed according to Algorithm 1. First, the algorithm applies pooling with a filter of size 1 and the same padding, which corresponds to the conventional mean squared error (MSE) calculation between the target template $\mathbf{K}$ and the output of the generator $\mathbf{Y}$. Note that the target template $\mathbf{K}$ is the sample taken from the extensive dataset consisting of both UEs and the corresponding optimal UAVs locations generated by the *core-set algorithm*. Then, by doubling the pooling size during each iteration, we provide important feedback to the generator about the spatial dislocation of the generated UAVs positions $\mathbf{Y}$ with respect to their optimal positions $\mathbf{K}$.

Finally, we formulate the optimization function of the cGAN with the proposed multilayer sum-pooling loss $\mathcal{L}_{pool}$ as follows

$$\arg \min_G \max_D \mathcal{L}_{cGAN}(G, D) + \lambda \mathcal{L}_{pool}(G) \quad (4)$$

### Deployment phase

The deployment phase is executed online to determine the UAVs positions in real time, respecting the UEs positions in the space. Here, we rely on the application of the reduced cGAN architecture, namely, its trained generator $G$ (see Figure 1(b)). By its nature, generator $G$ generates blurred images, resulting in the presence of a larger number of UAVs in the fleet than necessary. To tackle this issue, a correction algorithm is developed and included in the deployment phase.

*Correction algorithm for coping with UAV blurring.* The conventional application of a cGAN in the area of image processing does not require accurate image precision and often suffers from image blurring. This is not of major concern in the original image processing domain and could be suppressed by advanced image processing techniques. However, in our case, as a side effect, the blurriness of the output results in an increased number of UAVs slightly displaced from each other in the relatively sparse matrix $\mathbf{Y}$.

To tackle this challenge, we employed a correction mechanism to address this issue, which is designed as follows. The proposed correction algorithm takes the sparse matrix $\mathbf{Y}$, and to accelerate the computations, it transforms it into a scalable coordinate storage scheme represented by the set $Y$,[25] solely encompassing the space coordinates $[x, y]$ where UAVs are present. Note that the transformation $f$ of the sparse matrix to the compressed set $f : \mathbf{Y} \to Y$ allows for faster computations than that with regular sparse matrices. In the first step, the algorithm randomly takes a UAV and its position $[x, y]$ from $Y$. Then, it determines all other UAVs positioned in $Y$ in the perimeter of size $\epsilon$ and calculates the mutual distance. The selected UAV and neighboring UAVs within the perimeter $\epsilon$ are withdrawn from $Y$. The group of the dedicated UAVs is replaced by the new particular UAV stored in the new set $L$ with

**Algorithm 2.** Correction algorithm

1: Input: The raw output of generator G with UAVs positions in the spatial matrix $\mathbf{Y}^{n \times n}$
2: Input matrix transformation: $f : \mathbf{Y} \to Y$
3: Initialize empty set $L = \emptyset$ and threshold $\epsilon$
4: **while** Y is not empty **do**
5: Pop first position from Y, $y = pop(Y)$
6: Calculate distances to all remaining UAVs locations in Y: $D_i = dist(y, Y_i), \forall i \in <1, size(Y)>$
7: Pop all m UAVs for which the distance is lower than threshold $\epsilon$, $P = pop(Y_i, D_i < \epsilon)$
8: Calculate the average coordinates $[\Delta x, \Delta y]$ $\Delta x = \frac{\sum_{i=1}^{m} x_i}{m}, \forall x_i \in P, i \in <1, size(P)>$, $\Delta y = \frac{\sum_{i=1}^{m} y_i}{m}, \forall y_i \in P, i \in <1, size(P)>$
9: Insert new UAV with coordinates $[\Delta x, \Delta y]$ into L, $L = L \cup [\Delta x, \Delta y]$
10: Output: L

$[\Delta x, \Delta y]$ coordinates computed as the ensemble average of all mutual distances calculated in this time step. The output of the algorithm is represented by the coordinate storage scheme L that contains the final number of UAVs along with their corresponding coordinates in the space. In our case, the hyperparameter $\epsilon$ was tuned by a large set of computational experiments, with the aim of maximizing the cost function presented in equation (1). The detailed steps of the algorithm are summarized in Algorithm 2.

Figure 2 graphically illustrates the issue of UAV blurring (red (thin) circles) and the correction mechanism presented in Algorithm 2 applied on top of the blurred UAVs positions (black bold circles).

## Performance evaluation and discussion

### Simulation model and basic assumptions

The simulation model is defined in the form of a geometric disk coverage problem.[26] The objective of the problem is to cover a set of UEs in a target area by the minimum number of UAVs with a given radius of coverage.

The performance is evaluated from two viewpoints: coverage quality and computational efficiency. As the reference for the coverage quality, we use the core-set algorithm, which provides the optimal solution but is not computationally efficient. The computational efficiency is assessed versus the spiral algorithm, which is the current best practice among quasi-optimal solutions in terms of computational efficiency.[9]

### Complexity assessment

We focus our analysis on the complexity of the generator G and proposed routine to filter the number of UAVs at the output of generator G, thus leaving the complexity of the training procedure out of the investigation. Note that the training procedure is realized prior to the UAVs deployment (offline) and thus does not negatively affect the computational requirements of the proposed approach in practical realizations. Thus, in what follows, we use the term *proposed approach* to indicate the joint application of the generator function G and proposed correction algorithm (i.e. deployment phase). As the baseline scenario, we present the complexity of the core-set and spiral algorithms and the conventional K-means clustering algorithm.

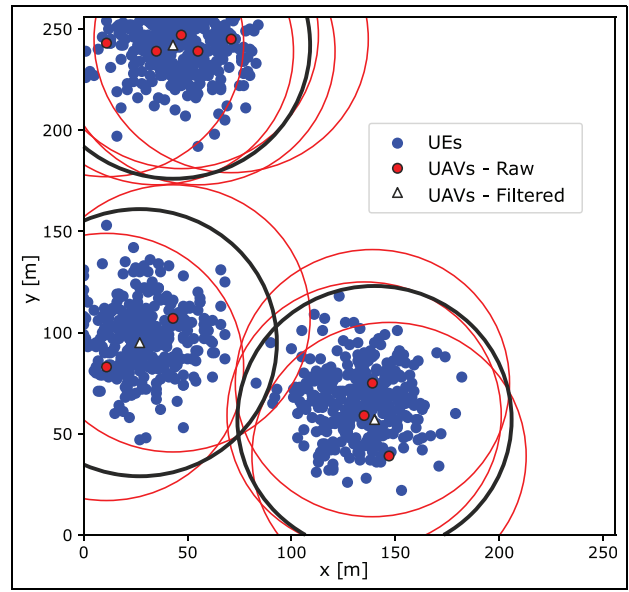

**Figure 2.** Deployment phase: directly predicted UAVs from generator G (red circles) and output of the correction routine (white triangles). R = 4 and number of UEs, p = 1000.

Let us denote the number of UAVs at the output of the generator G as y and number of filtered UAVs at the output of the correction mechanism as k. Based on massive numerical experiments, we can claim that in general, $y \sim 2.4k$. As indicated by the pseudocode in Algorithm 2, the routine resembles the typical array sorting procedure with a complexity of $\mathcal{O}(y^2)$. As $y \sim 2.4k$, we can approximate the complexity as $\mathcal{O}(k^2)$. Note that this is the worst-case scenario since the reduction of the sorting set generally leads to reduced number of executed inner steps. The time complexity of the analytical solution provided by the core-set algorithm is by definition $\mathcal{O}(p^k)$, where $p$ is the number of UEs in the area. The time complexity of the spiral algorithm is $\mathcal{O}(k^3)$, which is significantly higher than that of the proposed approach. Among the conventional well-known machine learning techniques, the K-means algorithm has a time complexity of $\mathcal{O}(p^{dk+1})$, where $d$ denotes the number of dimensions of the state space (in our case, $d = 2$).[27] While the time complexity of both the core-set and K-means algorithms depends on the number of UEs in the target area, for the spiral and proposed approaches, the time complexity is insensitive to the number of UEs. The time complexity of the spiral algorithm is cubic, while our approach runs quadratic in time as the worst-case scenario. As the downside of our approach, the training procedure must be executed prior to UAVs optimization deployment, which is not the case for the other investigated algorithms. A comparison of the time complexities of the given algorithms is shown in Table 1. Note that for the experiments, we used a workstation equipped with an AMD Ryzen Threadripper 1950X with 64 RAM memory, and the simulations were executed on an NVIDIA GeForce GTX 1080Ti 11 GB GPU. The simulator routines were programmed in Python ver. 3.6, running on MS Windows 10 Pro.

**Table 1.** Complexity requirements.

| Algorithm | Time c | Training | Solution |
|---|---|---|---|
| Proposed | $\mathcal{O}(k^2)$ | Yes | Heuristic |
| Core-sets | $\mathcal{O}(p^k)$ | No | Deterministic |
| K-means | $\mathcal{O}(p^{dk+1})$ | No | Heuristic |
| Spiral | $\mathcal{O}(k^3)$ | No | Deterministic |

### Coverage analysis

We define a coverage ratio metric $R = D/r$, where $D$ is a side length of the target square coverage area and $r$ is the radius of the coverage circle of a single UAV. We tested three coverage ratios $R = 2, 4, 6$ for two different numbers of UEs: 400 and 1000. The cGAN was trained only once for each $R$, and the same weights were transferred to predict UAVs positions for different number of UEs. Additionally, the correction parameter $\epsilon$ was adjusted during simulations to remove the effect of UAV blurring.

In Table 2, we can see the averaged simulation results of the proposed algorithm in comparison with the spiral and core-set algorithms. The statistics were averaged over the dataset with a size of 1000 samples (snapshots of various instantaneous UEs positions). We can appreciate the time complexity reduction of the

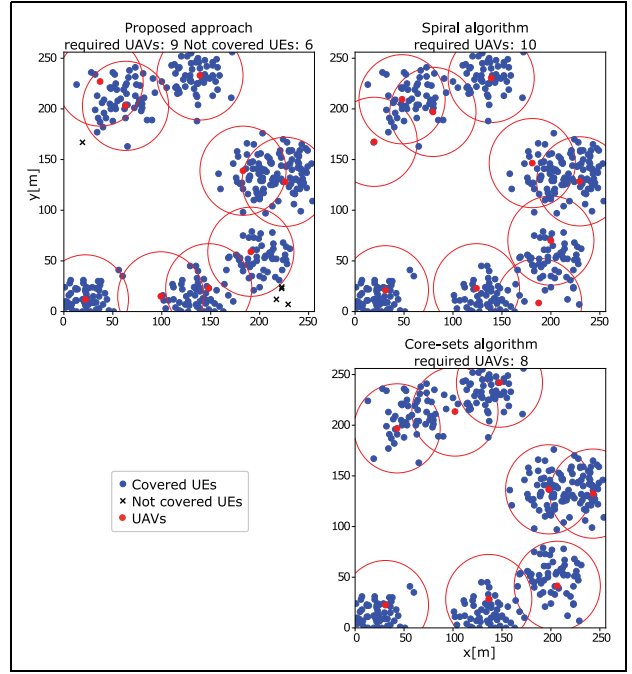

**Figure 3.** Computational comparison of the proposed approach, core-set algorithm, and spiral algorithm, $R = 6$, 1000 UEs. Note that in addition to the fastest processing capabilities, the proposed approach requires in general less UAVs than the quasi-optimal spiral algorithm.

proposed approach across all investigated scenarios. In addition, the number of required UAVs for the proposed approach is in general closer to the optimal core-set solution than to the spiral algorithm solution. Graphical representations of the coverage solutions of the proposed approach, spiral algorithm, and core-set algorithm are shown in Figure 3. As the downside of the proposed approach, we need to note that the coverage is not optimal and there are very few UEs that are not covered (also noted in Table 2). These are mostly located at the edges of the UEs clusters or are outliers not captured by the generator $G$. As long as our interest is to cover high-density clusters of UEs, we do not see this downside as a major problem. Obviously, these UEs can be easily covered by traditional ground infrastructure, that is, eNodeB network elements. Considering the growing demand for mobile applications and exponential growth in connected devices in smart cities, we can envision that the typical number of UEs in the foreseeable future will be an order of magnitude higher than those simulated in this article. Thus, we assume that the proposed approach with an asymptotic complexity of $\mathcal{O}(k^2)$ can be adopted for the deployment of aerial wireless communication systems, as well as for other UAV applications that are facing the disk coverage problem. *Finally, considering that the cGAN can be trained over any type of data (i.e. more*

**Table 2.** Time characteristics of each algorithm according to R and the number of UEs (left), corresponding coverage characteristics, and required average number of UAVs (right).

| Settings | | | Time (s) | | | Average coverage (%) (proposed approach) | Average number of UAVs | | |
|---|---|---|---|---|---|---|---|---|---|
| R | UEs | $\epsilon$ | Proposed approach | Spiral | Core-sets | | Proposed approach | Spiral | Core-sets |
| 2 | 400 | 160 | **0.07191** | 0.83652 | 0.16217 | 98.9 | 2 | 1.8 | 1.7 |
|   | 1000 | 160 | **0.08767** | 5.49683 | 0.71645 | 99.8 | 1.9 | 2 | 1.9 |
| 4 | 400 | 60 | **0.05183** | 0.45599 | 5.12239 | 98.9 | 4.1 | 4.3 | 3.7 |
|   | 1000 | 60 | **0.05295** | 2.80836 | 38.92758 | 99.7 | 3.9 | 4.4 | 3.7 |
| 6 | 400 | 40 | **0.0804** | 0.32621 | 264.4874 | 97.6 | 7.1 | 7.6 | 6.5 |
|   | 1000 | 40 | **0.0967** | 1.41399 | 8493.39611 | 99.1 | 7.9 | 8.5 | 6.8 |

UE: user equipment; UAV: unmanned aerial vehicle.
For the spiral and core-set algorithms, by definition, the coverage is 100%. In comparison, the proposed approach provides a tremendous time complexity improvement accompanied by only a slight performance degradation in terms of coverage (see also Figure 3). Note that the time complexity requirements for the spiral and core-set algorithms significantly increase with increasing number of UEs, while that of the proposed approach remains almost constant. The best achieved time for a given setting is highlighted in bold.

*complex datasets), the proposed algorithm can be easily extended to complex coverage optimization based on multiple wireless network performance criteria, such as achievable throughput and area spectral efficiency*. Thus, the solution based on the application of a cGAN proposed in this article appears to be feasible, opening further promising research directions in this area.

## Conclusion

To improve the deployment of moving cells based on UAVs, we have proposed a new computationally efficient algorithm for coverage optimization. The proposed approach is based on a cGAN, with a unique multilayer sum-pooling loss function that allows convergence to the quasi-optimal coverage solution in the presence of high-density clusters of UEs. Simulations have been conducted to assess the performance of the proposed system versus the optimal solution provided by the core-set algorithm and the most computationally efficient quasi-optimal solution provided by the spiral algorithm. The results show that the proposed approach provides similar results to those provided by other state-of-the-art algorithms while being an order of magnitude better in terms of computational time. In future research, we will enhance the coverage optimization by using deep RL with the reward function determined by the cGAN algorithm based on multiple network performance indicators.

## Declaration of conflicting interests

The author(s) declared no potential conflicts of interest with respect to the research, authorship, and/or publication of this article.

## Funding

The author(s) disclosed receipt of the following financial support for the research, authorship, and/or publication of this article: This work was supported by the Slovak Research and Development Agency, project numbers APVV-18-0214 and APVV-15-0055, by the Scientific Grant Agency of the Ministry of Education, Science, Research and Sport of the Slovak Republic under contract 1/0268/19, by the National Natural Science Foundation of China (No. 61962036), and by the Ukrainian government project No. 0120U100674, "Designing the novel decentralized mobile network based on blockchain architecture and artificial intelligence for 5G/6G development in Ukraine."

## ORCID iD

Juraj Gazda  https://orcid.org/0000-0002-7334-9540

## References

1. Ritchie H. Google mobility trends: how has the pandemic changed the movement of people around the world? https://ourworldindata.org/covid-mobility-trends (accessed 8 December 2021).
2. Fotouhi A, Qiang H, Ding M, et al. Survey on UAV cellular communications: practical aspects, standardization advancements, regulation, and security challenges. *IEEE Commun Surv Tutor* 2019; 21(4): 3417–3442.
3. Mozaffari M, Saad W, Bennis M, et al. A tutorial on UAVs for wireless networks: applications, challenges, and open problems. *IEEE Commun Surv Tutor* 2019; 21(3): 2334–2360.
4. Gupta L, Jain R and Vaszkun G. Survey of important issues in UAV communication networks. *IEEE Commun Surv Tutor* 2015; 18(2): 1123–1152.
5. Lai C, Chen C and Wang L. On-demand density-aware UAV base station 3D placement for arbitrarily

distributed users with guaranteed data rates. *IEEE Wirel Commun Le* 2019; 8(3): 913–916.
6. Pan C, Yi J, Yin C, et al. Joint 3D UAV placement and resource allocation in software-defined cellular networks with wireless backhaul. *IEEE Access* 2019; 7: 104279–104293.
7. Bādoiu M, Har-Peled S and Indyk P. Approximate clustering via core-sets. In: *Proceedings of the thirty-fourth annual ACM symposium on theory of computing*, Montreal, QC, Canada, 19–21 May 2002, pp. 250–257. New York: Association for Computing Machinery.
8. Fayed HA and Atiya AF. A mixed breadth-depth first strategy for the branch and bound tree of Euclidean k-center problems. *Comput Optim Appl* 2013; 54(3): 675–703.
9. Lyu J, Zeng Y, Zhang R, et al. Placement optimization of UAV-mounted mobile base stations. *IEEE Commun Lett* 2016; 21(3): 604–607.
10. Qu Y, Zhang J, Li R, et al. Generative adversarial networks enhanced location privacy in 5G networks. *Sci China Inf Sci* 2020; 63(12): 1–12.
11. Hughes B, Bothe S, Farooq H, et al. Generative adversarial learning for machine learning empowered self organizing 5G networks. In: *2019 international conference on computing, networking and communications (ICNC)*, Honolulu, HI, 18–21 February 2019, pp. 282–286. New York: IEEE.
12. Wu B, Liu L, Yang Y, et al. Using improved conditional generative adversarial networks to detect social bots on Twitter. *IEEE Access* 2020; 8: 36664–36680.
13. Haghighat MH and Li J. Intrusion detection system using voting-based neural network. *Tsinghua Sci Technol* 2021; 26(4): 484–495.
14. Plachy J, Becvar Z, Mach P, et al. Joint positioning of flying base stations and association of users: evolutionary-based approach. *IEEE Access* 2019; 7: 11454–11463.
15. Plachy J and Becvar Z. Energy efficient positioning of flying base stations via Coulomb's law. In: *2020 IEEE globecom workshops (GC Wkshps)*, Taipei, Taiwan, 7–11 December 2020, pp. 1–6. New York: IEEE.
16. Gonzalez V, Monje C, Garrido S, et al. Coverage mission for UAVs using differential evolution and fast marching square methods. *IEEE Aero El Sys Mag* 2020; 35(2): 18–29.
17. Horváth D, Gazda J, Šlapak E, et al. Evolutionary coverage optimization for a self-organizing UAV-based wireless communication system. *IEEE Access* 2021; 9: 145066–145082.
18. Wang L, Wang K, Pan C, et al. Multi-agent deep reinforcement learning-based trajectory planning for multi-UAV assisted mobile edge computing. *IEEE Trans Cogn Commun Netw* 2021; 7(1): 73–84.
19. Cui J, Ding Z, Deng Y, et al. Adaptive UAV-trajectory optimization under quality of service constraints: a model-free solution. *IEEE Access* 2020; 8: 112253–112265.
20. Khan MF and Yau KLA. Route selection in 5G-based flying ad-hoc networks using reinforcement learning. In: *2020 10th IEEE international conference on control system, computing and engineering (ICCSCE)*, Penang, Malaysia, 21–22 August 2020, pp. 23–28. New York: IEEE.
21. Liu X, Zhou Q, Cheng CT, et al. A greedy-model-based reinforcement learning algorithm for Beyond-5G cooperative data collection. *Phys Commun* 2021; 50: 101496.
22. Mirza M and Osindero S. *Conditional generative adversarial nets*. arXiv preprint arXiv:1411.1784, 2014.
23. Goodfellow IJ, Pouget-Abadie J, Mirza M, et al. *Generative adversarial networks*. arXiv preprint arXiv:1406.2661, 2014.
24. Isola P, Zhu JY, Zhou T, et al. Image-to-image translation with conditional adversarial networks. In: *Proceedings of the IEEE conference on computer vision and pattern recognition*, Honolulu, HI, USA, 21–26 July 2017, pp. 1125–1134. New York: IEEE.
25. Hossain S and Steihaug T. Sparse matrix computations with application to solve system of nonlinear equations. *Wiley Interdiscip Rev Comput Stat* 2013; 5(5): 372–386.
26. Srinivas A, Zussman G and Modiano E. Construction and maintenance of wireless mobile backbone networks. *IEEE ACM T Network* 2009; 17(1): 239–252.
27. Shewchuk JR. Delaunay refinement algorithms for triangular mesh generation. *Comput Geo* 2002; 22(1–3): 21–74.